\documentclass{article}

\usepackage{arxiv}
\usepackage[outdir=./]{epstopdf}

\usepackage[utf8]{inputenc} 
\usepackage[T1]{fontenc}    
\usepackage{hyperref}       
\usepackage{url}            
\usepackage{booktabs}       
\usepackage{amsfonts}       
\usepackage{nicefrac}       
\usepackage{microtype}      
\usepackage{lipsum}
\usepackage{amsmath,graphicx}
\usepackage{float}
\usepackage{booktabs}
\usepackage{url}
\usepackage{gensymb}

\usepackage{adjustbox}
\usepackage{graphicx}
\usepackage{tikz}

\title{Level Three Synthetic Fingerprint Generation}

\author{
André Brasil Vieira Wyzykowski, Mauricio Pamplona Segundo, Rubisley de Paula Lemes \thanks{This research was funded by CNPq (\url{http://cnpq.br/}). The Titan Xp used for this research was donated by the NVIDIA Corporation.}
 \\
  Intelligent Vision Research Lab\\
  Federal University of Bahia, Brazil
}

\begin{document}
\maketitle

\begin{abstract}
Today's legal restrictions that protect the privacy of biometric data are hampering fingerprint recognition researches. For instance, all high-resolution fingerprint databases ceased to be publicly available. To address this problem, we present a novel hybrid approach to synthesize realistic, high-resolution fingerprints. First, we improved Anguli, a handcrafted fingerprint generator, to obtain dynamic ridge maps with sweat pores and scratches. Then, we trained a CycleGAN to transform these maps into realistic fingerprints. Unlike other CNN-based works, we can generate several images for the same identity. We used our approach to create a synthetic database with 7400 images in an attempt to propel further studies in this field without raising legal issues. We included sweat pore annotations in 740 images to encourage research developments in pore detection. In our experiments, we employed two fingerprint matching approaches to confirm that real and synthetic databases have similar performance. We conducted a human perception analysis where sixty volunteers could hardly differ between real and synthesized fingerprints. Given that we also favorably compare our results with the most advanced works in the literature, our experimentation suggests that our approach is the new state-of-the-art.
\end{abstract}


\section{Introduction}
\label{sec:intro}
Fingerprint recognition is widely studied thanks to its compliance with the core premises of biometrics: permanence and distinctiveness~\cite{Maltoni2009, jain2010biometrics}. It is possible to analyze the ridge patterns on fingerprints at different scales and resolutions. These patterns can be classified as Level 1 (L1 - global patterns, such as ridge orientation maps, and fingerprint classes), Level 2 (L2 - local patterns, such as minutiae) and Level 3 (L3 - fine details, such as sweat pores, incipient ridges and dots). 

With the development of high-resolution sensors able to capture L3 fingerprint images, researchers saw an opportunity to devise new and more accurate recognition approaches by using the most exploited L3 feature in the literature, the presence of sweat pores~\cite{7301328, DBLP:journals/corr/abs-1905-06981, DBLP:journals/corr/abs-1905-11472}. Besides, L3-based approaches improve security by hindering spoof attempts~\cite{articleEspinoza, inproceedingsMarcialis, inproceedingsSilva}.

Despite the recent improvements brought to the fingerprint recognition research area, L3 fingerprint databases are being discontinued. For instance, databases such as the NIST Special Database 30~\cite{nist30} and the Hong Kong Polytechnic University High Resolution Fingerprint database (PolyU)~\cite{10.1007/978-3-642-01793-3_61} are no longer publicly available. More recently, Anand and Kanhangad~\cite{an2019crosssensor, Anand2019} created L3 databases for their recognition experiments. However, until the present moment, these databases were not released. The main reason for that is the existence of legal restrictions protecting the privacy of biometric data, which hinders the evolution of fingerprint recognition research.

Creating a synthetic database is a valid alternative to solve this problem. Synthetic generation techniques are widely exploited in several areas, such as optical flow computation \cite{butler2012naturalistic}, indoor robot navigation \cite{wu2018building}, and autonomous driving \cite{wrenninge2018synscapes}. The same is true for the fingerprint recognition field.

Earlier works, such as SFinGe~\cite{article_sfinge} and Anguli\cite{ansari2011generation}, developed handcrafted frameworks for fingerprint generation based on the knowledge of an expert. Their design allows reasonable control over the identity of the output images, but lack realism, especially when considering L3 features. Besides, SFinGe restricts the generation of large datasets, difficulting the generation of multiple instances of a single identity. Finger-GAN ~\cite{minaee2018fingergan} and Cao and Jain's approach \cite{8411200} employed Generative Adversarial Networks (GAN)
~\cite{goodfellow2014generative} to learn how to generate realistic fingerprints from a training set of real examples. However, both of these works present many unnatural artifacts and cannot generate multiple images for the same identity.

We propose a novel approach to create realistic, high-resolution synthetic fingerprints while maintaining control over the identity of the output images. Our goal is to foster further studies in this field without raising legal issues that come with real biometric data. Our contributions are:

\begin{enumerate}
    \item A novel hybrid fingerprint generation approach that combines a handcrafted identity generator and a learned texturizer to achieve realistic results and generate multiple images of a single identity. A visual comparison to existing approaches shows that our results are the new state-of-the-art. Also, a perception experiment shows that humans can hardly differ between real and our synthetic images.
    \item A public database\footnote{\url{https://andrewyzy.github.io/L3-SF/}} of L3 synthetic fingerprint images with five subsets of 148
    identities, with 10 samples per identity, totaling 7400 fingerprint images. Also, we include sweat pore annotations for 740 images to assist in pore detection research \cite{de2014dynamic}. This is the largest publicly available database with L3 fingerprints nowadays.
\end{enumerate}

\section{Synthetic Fingerprint Generation}
\label{HIGHRESOLUTION}

\begin{figure}[H]
   \centering
            \begin{tabular}{cc}
            \includegraphics[scale=0.18]{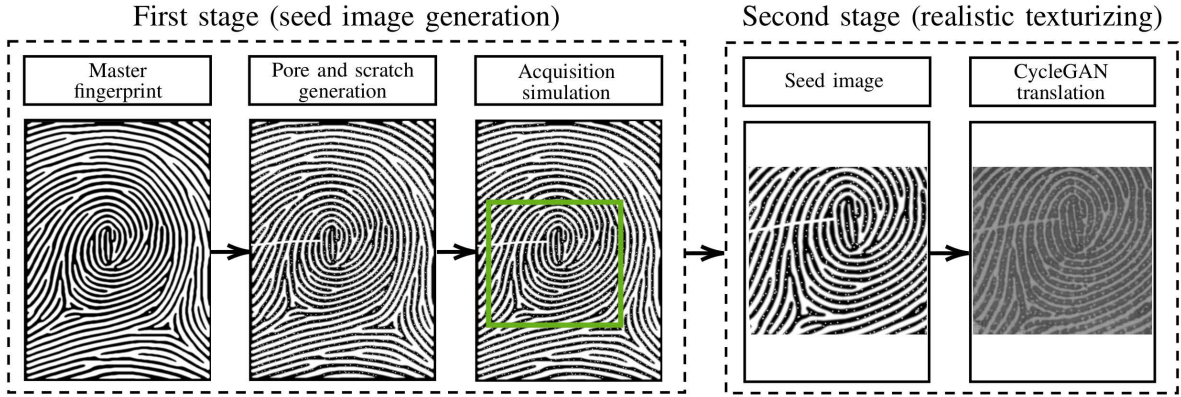} 
            \end{tabular}

\caption{Flowchart with the steps to create a high-resolution synthetic fingerprint using the proposed approach.}
\label{Desenvolvimento_FLUXO}
\end{figure}

To generate high-resolution synthetic fingerprints, we split our approach into two stages. The first stage concerns procedures required to create multiple instances of fingerprints, which we call \textbf{seed images}. The second stage consists of using CycleGAN~\cite{ZhuPIE17} to translate seed images into realistic L3 fingerprints. Figure~\ref{Desenvolvimento_FLUXO} summarizes the workflow required to create a high-resolution synthetic fingerprint using the proposed method.

The first stage consists of the following processes:
\begin{enumerate}
    \item \textbf{Master fingerprint generation:} we extended Anguli~\cite{ansari2011generation}, an open-source implementation of SFinGe~\cite{article_sfinge}, to create fingerprint ridge maps with random ridge-flow frequency. Also, we segment ridges and dynamically change their thicknesses. We call the resulting images master fingerprints (Section \ref{Masterfingerprint}).
    \item \textbf{Pore and scratch generation:} we add pores and scratches to each master fingerprint following a distribution learned from real images to obtain L3 master fingerprints (Section \ref{poreGENERATOR}).
    \item \textbf{Fingerprint acquisition simulation:} to simulate acquisition, we randomly cut the L3 master fingerprints following a distribution of the displacement among images of the same person in a real database, thus creating different instances for each identity. These instances are called seed images (Section \ref{seedImagesGeneration}).
\end{enumerate}

The second stage consists of performing CycleGAN translation to transform any seed image into a realistic L3 fingerprint. The model employed in this step is obtained through the following processes:

\begin{enumerate}
    \item \textbf{Training set generation:} to train CycleGAN, we create several seed images following the steps of the first stage of our approach.  Data augmentation is then applied to real and seed images to compose the training set (Section \ref{TrainingStep}).
    \item \textbf{CycleGAN training:} uses the training set to create a CNN model that translates seed images to high-resolution fingerprints (Section \ref{cyclegantranslation}).
\end{enumerate}

\subsection{Master fingerprint generation}
\label{Masterfingerprint}

\begin{figure}[H]
   \centering
            \begin{tabular}{cc}
            \includegraphics[scale=0.25]{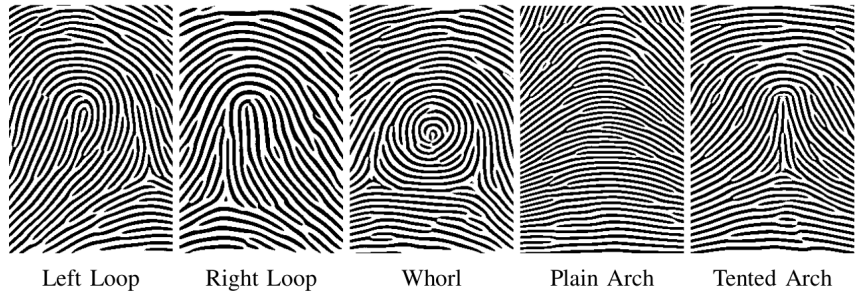} 
            \end{tabular}

    \caption{Five fingerprint class patterns created by our modified version of Anguli.}
    \label{5tiposAnguli} 
\end{figure}

We use Anguli~\cite{ansari2011generation} to create the ridge map that composes a master fingerprint within one of the following classes: Whorl, Right Loop, Left Loop, Plain Arch, and Tented Arch. Figure~\ref{5tiposAnguli} illustrates these classes. When creating a new identity, it is necessary to follow the proportion of these classes in a real population. Martijn van Mensvoort~\cite{martijnvanmensvoort2015} gathered fingerprint distributions from 32 countries through a compilation of 28 published articles. Each country has its peculiarities and proportions, so we decided to use the global mean distribution. The mean distribution for 32 countries is: Whorl: 41\%, Right Loop: 50\%, Left Loop: 3\%, Arch: 6\%. Since Anguli can create two arch fingerprint types,  we used the ratio provided by Wang~\cite{Pattern2011}: Plain Arch: 72.22\% and Tented Arch: 27.77\%.

The Gabor filter plays an important role when generating ridges using Anguli. By changing its scale, we introduce variations to the ridge frequency in synthetic fingerprints. We changed Anguli to randomly modify the scale of the Gabor filter by up to 20\%. To exemplify this effect, we show images in Figure~\ref{5tiposAnguli} with different frequencies. As can be observed, the first image frequency is much higher than the second one.
 
To segment ridges, we upscale the modified Anguli images, which have 275$\times$400 pixels, by a factor of 3 with FSRCNN \cite{dong2016accelerating} and smooth the result by applying a 3x3 mean filter. Then, we skeletonize those images using Zhang and Suen's thinning algorithm~\cite{Zhang:1984:FPA:357994.358023} and split continuous segments of pixels as individual ridges.

\begin{figure}[H]
   \centering
            \begin{tabular}{cc}
            \includegraphics[scale=0.2]{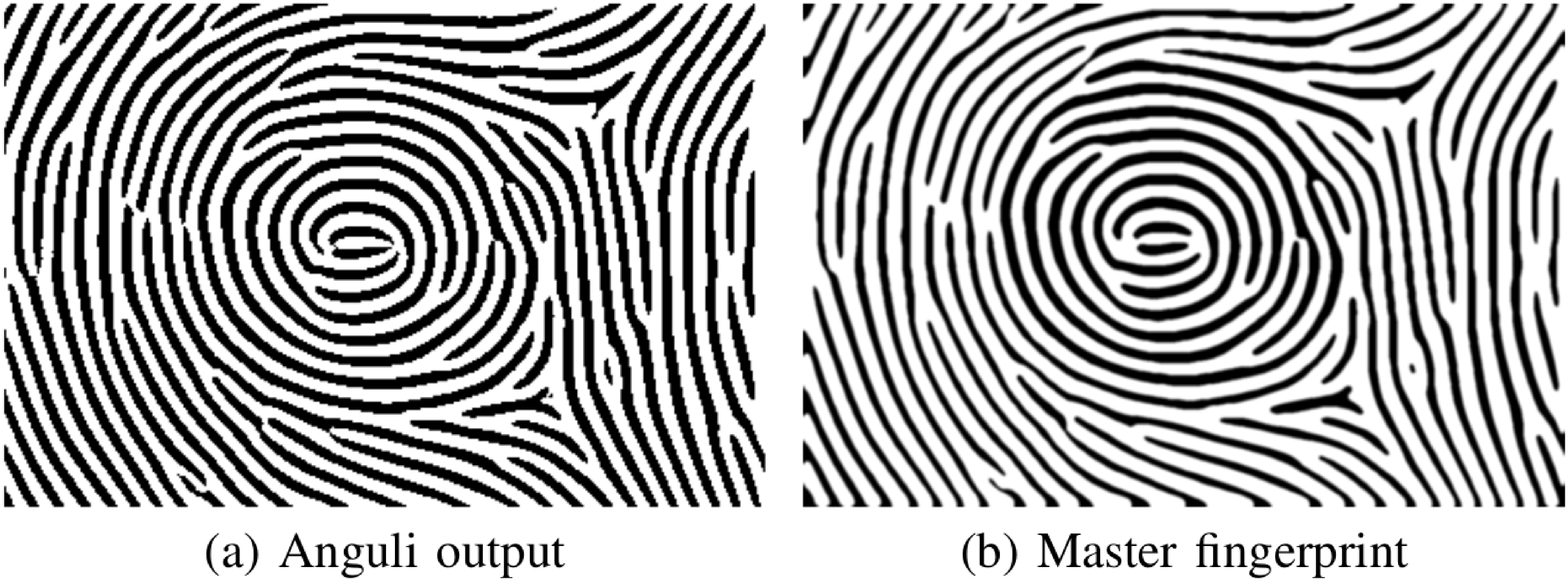} 
            \end{tabular}

    \caption{Example of (a) an output of our modified version of Anguli and (b) its respective master fingerprint after ridge thickness sinusoidal perturbation. Note the ridge thickness variability in the master fingerprint, where the ridges are thicker on the center compared to the borders. Also, note our aliased result in comparison to Anguli's pixelated ridges.}
    \label{thickness} 
\end{figure}

To create a higher variability in the ridges' thicknesses, we dynamically change ridge thickness based on the sine function, generating a smooth transition among neighboring ridges. We iteratively calculate $w_i = | 3 \times \sin{t}|$, where $w_i$ is the width of the i-th ridge and $t$ is a counter starting at a random value for each image. After processing a ridge, $t$ is incremented by 0.1. Ridges are processed from left to right, top to bottom. This approach avoids abrupt changes in thickness among neighboring ridges and also adds a stochastic factor to the thickness generation. The outcome is a \textbf{master fingerprint} with 825$\times$1200 pixels, as illustrated in Figure~\ref{thickness}.

\subsection{Pore and scratch generation}
\label{poreGENERATOR}

This step consists in marking where the pores will be placed on the ridges and applying the scratches on the images. To measure the distance distribution from one pore to another, we use a pore-based ridge reconstruction approach~\cite{7301328}. Given a training set of real fingerprint images, we compute the average distance and the standard deviation among neighboring pores as our reference distribution.

\begin{figure}[H]
   \centering
            \begin{tabular}{cc}
            \includegraphics[scale=0.2]{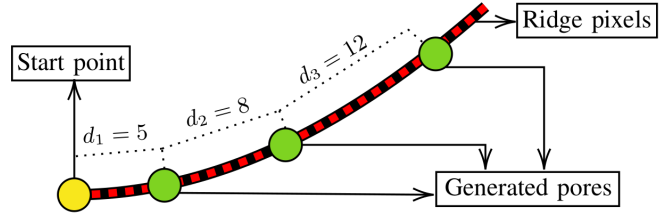} 
            \end{tabular}

  \caption{Illustration of the pore generation process. Red squares represent a sequence of pixels in a ridge and the green circles are the generated pores. Given a start point, pore distances are sampled from the reference distribution. In this example, new pores are created with distances $ d_1 = 5$,  $ d_2 = 8$ and $ d_3 = 12$.}
  \label{criacaoPorosExemplo}
\end{figure}

To add pores, we utilize the segmented ridges generated in Section \ref{Masterfingerprint}. Starting from the beginning of a ridge, we iteratively sample distances $d_i$ from the reference distribution. We follow the ridge pixels to add the i-th pore $d_i$ away from the previous pore until the end of the ridge is reached. This process of creating pores is illustrated in Figure~\ref{criacaoPorosExemplo}.

\begin{figure}[H]
   \centering
            \begin{tabular}{cc}
            \includegraphics[scale=0.2]{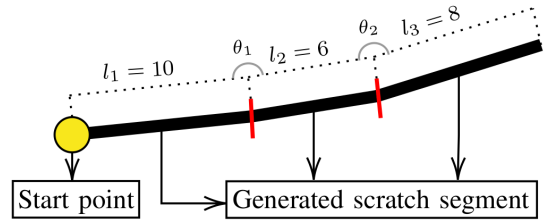} 
            \end{tabular}

  \caption{Illustration of the scratch generation process. Given a random start point, we draw $n$ line segments with random length and random angle between them. In this example, there are 3 lines segments with lengths $ l_1 = 10$,  $ l_2 = 6$ and $ l_3 = 8$.}
  
  \label{criacaoLinhasExemplo}
\end{figure}

To create scratches, we count the number of scratches in each fingerprint on a real database. With these values, we used the normalized cumulative density function, and we choose the number of scratches based on a uniform random number. Starting on a random point, for each scratch, we draw $n$ consecutive line segments, where $n$ is a random value between 1 and 4. The line segments have a random length with a maximum value of 150 pixels and a random angle ($-15\degree  \leq \theta \leq 15\degree$) between them. This process of creating scratches is illustrated in Figure~\ref{criacaoLinhasExemplo}.

\begin{figure}[H]
   \centering
            \begin{tabular}{cc}
            \includegraphics[scale=0.2]{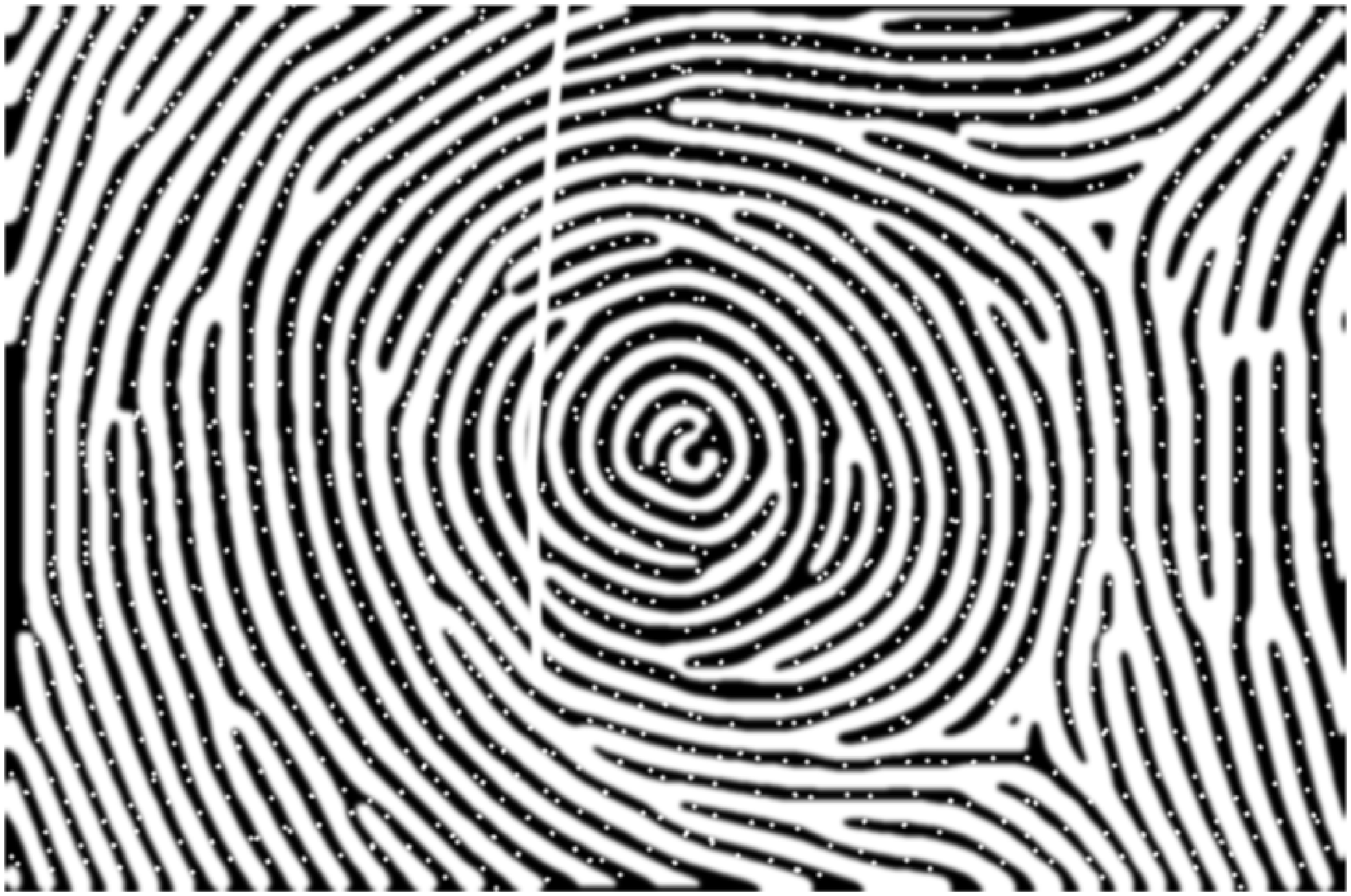} 
            \end{tabular}

    \caption{An L3 master fingerprint. Note the generated pores and scratch.}
    \label{l3exemplo} 
\end{figure}

Figure~\ref{l3exemplo} shows an \textbf{L3 master fingerprint}, the outcome of this step.

\subsection{Fingerprint acquisition simulation} 
\label{seedImagesGeneration}

Different finger positions on the image sensor produce distinct rotations and shifts on fingerprint images. Our approach aims to simulate these acquisition variations.

To measure rotation and shift variations between fingerprints of the same person from a training set, we extract SIFT ~\cite{Lowe:1999:ORL:850924.851523,Lowe2004} and ORB \cite{Rublee:2011:OEA:2355573.2356268} keypoints, perform an affine alignment \cite{yu2011asift} for each keypoint set separately, and select the one with the highest number of inliers. After, we use the RANSAC algorithm~\cite{Fischler:1981:RSC:358669.358692} to obtain a rigid transformation. 

We assume that the center of a L3 master fingerprint is the average of the center of its samples aligned to each other in the same coordinate system. With this assumption, the average shift of the samples in relation to the L3 master fingerprint center is zero in each axis, with standard deviations $\sigma_x$ and $\sigma_y$ independent from each other.

However, we cannot measure these displacements in a real dataset, as we have samples but do not have master fingerprints. What we can observe is the difference between two samples. With $x_i$ being the x-coordinate of the center of the i-th image, the expected difference between the center of two samples $i$ and $j$, aligned to the same coordinate system, is given by:

 \begin{equation*}\label{formula1}
  \begin{aligned}
\mathbb{E}[(x_i - x_j)^2]
&= \mathbb{E}\big[\mathcal{N}(0, {\sigma_x}^2) + \mathcal{N}(0, {\sigma_x}^2) \big] \\
&= \mathbb{E}\big[\mathcal{N}(0, 2{\sigma_x}^2)\big]
 \end{aligned}
\end{equation*}

Therefore, if we measure the average square difference $\overline{d}_x$ between pairs of samples from the same person, $\sigma_x$ can be estimated as:

 \begin{equation*}\label{formula2}
    \sigma_x = \sqrt{\frac{\overline{d}_x}{2}}
\end{equation*}

\begin{figure}[H]
   \centering
            \begin{tabular}{cc}
            \includegraphics[scale=0.2]{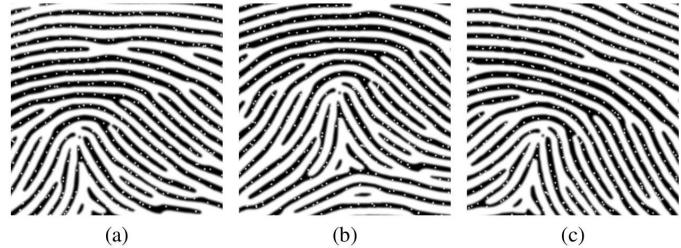} 
            \end{tabular}

    \caption{Seed images generated from a single L3 master fingerprint, presenting distinct shifts and rotations.}
    \label{seed_images_exemplos} 
\end{figure}

The same can be done independently for $\sigma_y$ and $\sigma_\theta$ (rotation). To create a seed image, we sample a random transformation from the normal distribution using $\sigma_x$, $\sigma_y$, and $\sigma_\theta$. After that, we rotate and translate the L3 master fingerprint before cropping the center region of size 512$\times$512 from the resulting image. Examples of seed images from the same L3 master fingerprint are shown in Figure~\ref{seed_images_exemplos}.

\begin{figure}[H]
   \centering
            \begin{tabular}{cc}
            \includegraphics[scale=0.15]{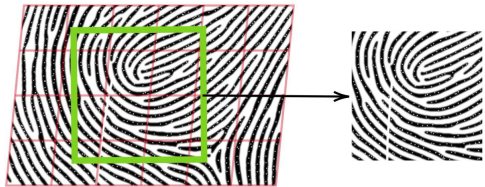} 
            \end{tabular}

  \caption{Illustration of an affine transformation over an L3 master fingerprint.}
  \label{AffineIlustation}
\end{figure}

Before cropping, we also perform a random affine transformation on L3 master fingerprints. Assuming that $(X, V, k)$ and $(Z, W, k)$ are two affine spaces, where $X$ and $Z$ are point sets, we generate a random $\gamma$ value between -10 and 10, summing $\gamma$ to $V$ and $W$ (vector spaces over the field $k$). This is a way to simulate non-rigid deformations on fingerprints, which occur in real images due to distinct finger pressures during acquisition or optical distortions caused by the acquisition sensor. Figure~\ref{AffineIlustation} illustrates this process. 

Finally, we included a method to randomly drop some pores in different seed images from a single identity. This is a way of simulating the process of perspiration. In this work, we used a pore dropout rate of 3\%.

\subsection{CycleGAN-based domain translation}
\label{cyclegantranslation}

To generate realistic fingerprint images, we learn to map the seed image domain into the real image domain of the chosen training database using CycleGAN~\cite{ZhuPIE17}.

CycleGAN is a viable solution for the task of translating two different domains as it does not require direct pairing between the training instances. Thus, our seed images do not need to be perfectly aligned to a real image in the training set.

When creating seed images for training, we seek to balance the number of real and synthetic samples. To increase the number of real training samples, besides performing horizontal flips, we take full real fingerprint images and apply the same acquisition simulation described in Section \ref{seedImagesGeneration} (except pore dropout). Section \ref{TrainingStep} describes how we create the training set for this work.

We use the original CycleGAN architecture with 13 residual blocks \cite{he2016deep} and input size 256$\times$256 (seed images are resized to these dimensions). Besides CycleGAN's cycle consistency loss, we use the identity mapping loss \cite{taigman2016unsupervised} as it contains a regularizing component that encourages the generator to map samples from the real fingerprint domain to themselves. We train our model using the Adam optimizer \cite{kingma2014adam} with a learning rate of 0.0002 for 3 epochs.

At the beginning of the training, the weight for the identity mapping loss is 0. We iteratively increase the weight up to $0.7 \times \lambda$, where $\lambda$ is the weight for the cycle consistency ($\lambda = 10$ in this work). We did this because CycleGAN tends to lose the master fingerprint identity, changing the ridge flow and the location of the minutiae. 

\begin{figure}[H]
   \centering
            \begin{tabular}{cc}
            \includegraphics[scale=0.15]{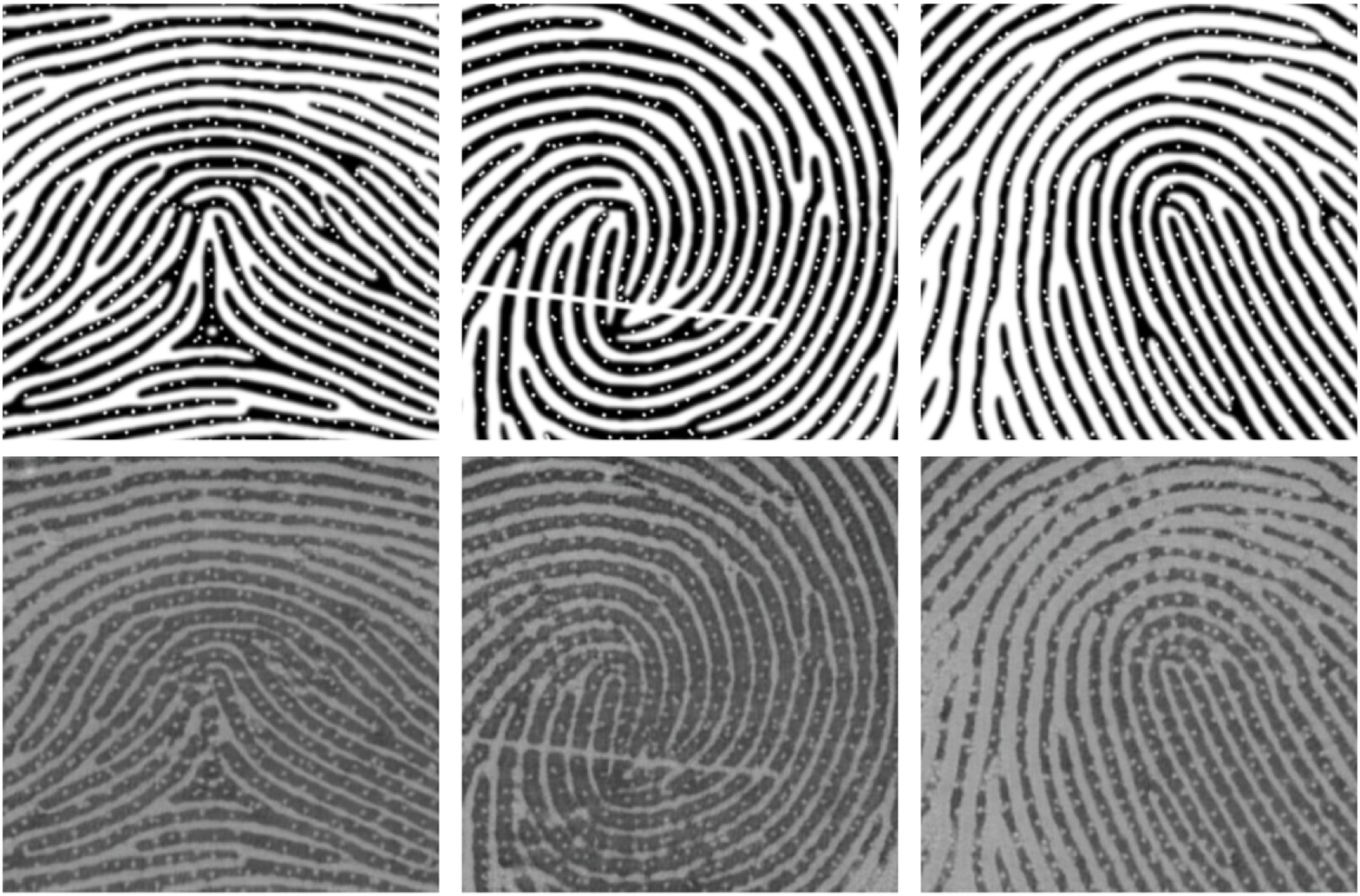} 
            \end{tabular}

  \caption{CycleGAN inference: seed images (top) and their respective results (bottom).}
  \label{exemplosSaidaCyclegan}
\end{figure}

After training, the outcome is a model that can translate any seed image into a realistic fingerprint, even if it was not seen during training. Examples of the inference using CycleGAN are presented in Figure~\ref{exemplosSaidaCyclegan}.

\section{Synthetic Database Creation}
\label{DATABASE}

We decided to replicate the structure of an existing database to be able to perform a comparison in terms of image realism and fingerprint recognition performance. To this end, we used the PolyU database~\cite{10.1007/978-3-642-01793-3_61}, which is divided in two subsets: DBI and DBII. DBI contains 320$\times$240 images of cropped fingerprints, while DBII contains 640$\times$480 images of full fingerprints. Both have 148 different identities, and each identity has 10 images acquired in two sessions (five per session). We use DBII for training purposes, as we need full fingerprints to perform the augmentation mentioned in Section \ref{cyclegantranslation}. More details on how we create the training set for CycleGAN are given in Section \ref{TrainingStep}. We then use the obtained model to create synthetic datasets with the same structure as DBI, as described in Section \ref{inferencestep}.

\subsection{CycleGAN training set}
\label{TrainingStep}

We created 296 L3 master fingerprints for training, which proved to be sufficient to map synthetic images to the real domain with CycleGAN.

We then flip these master fingerprints horizontally and perform the acquisition simulation described in Section \ref{seedImagesGeneration} to create 5920 \textbf{seed images}. We create additional 5920 seed images by repeating the acquisition simulation with a larger cropping larger area (825$\times$825 pixels) to cope with fingerprints with higher ridge frequency. Finally, we apply random elastic deformations~\cite{1227801} to the set of seed images to create 11840 distorted images, totaling 23680 training seeds.

For \textbf{real images}, we use the DBII subset of the PolyU database, which contains 1480 full fingerprint images. First, we flipped these images horizontally. After, we performed the acquisition simulation to create 10 samples per image, totaling 29600 real training samples. Table~\ref{augmentation} summarizes the training images and the augmentation operations.


\begin{table}[H]
\centering
    \caption{Training set images.}
\begin{tabular}{@{}lcc@{}}
\toprule
 & \textbf{Synthetic} & \textbf{Real} \\ \midrule
Initial images & 256 (master fingerprint) & 1480 \\
Flip horizontally & 592 (master fingerprint) & 2960 \\
\begin{tabular}[c]{@{}l@{}}Shift, crop and rotation (10 variations)\end{tabular} & 5920 (seed image) & 29600 \\
Higher ridge frequency cut & 11840 (seed image) & - \\
Elastic deformation~\cite{1227801} & 23680 (seed image) & - \\ \midrule
\textbf{Total images} & \textbf{23680} & \textbf{29600} \\ \bottomrule

\end{tabular}

  \label{augmentation}
\end{table}

\subsection{Inference for database generation}
\label{inferencestep}

We created five replicas of the PolyU DBI subset to establish a confidence interval in our recognition experiments. To do so, we generated 5 sets of 148 master fingerprints. After simulating fingerprint acquisition, we end up with 1480 seed images per set, totaling 7400 images for inference.

We then transform these seed images into real images using our CycleGAN model. To replicate the resolution and aspect ratio of PolyU DBI, we upscale the inferred images (256$\times$256) by a factor of 2 using FSRCNN \cite{dong2016accelerating}, crop the center region of size 512$\times$384 from the resized images (512$\times$512), and resize the crops to the PolyU DBI resolution (320$\times$240). These images, split into five subsets, compose our L3 synthetic fingerprint (L3-SF) database.

\section{Experimental results}
\label{sec:humanperception}

Section~\ref{sec:Visualanalysis} presents a visual analysis of the L3-SF database, including a visual comparison with other methods. Section~\ref{sec:Experiments} presents a fingerprint recognition analysis using both PolyU DBI and L3-SF databases. A human perception experiment involving 60 volunteers is reported in Section~\ref{sec:humanperception}.

 \subsection{Visual analysis}
\label{sec:Visualanalysis}

\begin{figure}[H]
   \centering
            \begin{tabular}{cc}
            \includegraphics[scale=0.2]{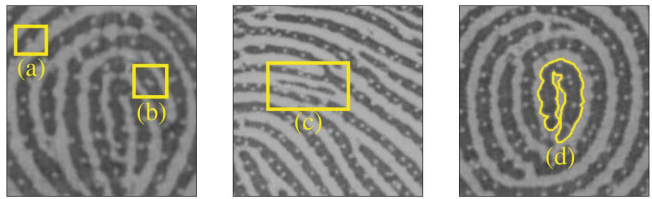} 
            \end{tabular}

    \caption{(a) and (b) are open and closed pores in an image of the L3-SF database. (c) shows an incipient ridge. (d) highlights an unique ridge shape.}
    \label{poro_open} 
\end{figure}

 We visually inspected our results to evaluate our method's ability to create realistic high-resolution fingerprint images. We observed that our method creates pores at the indicated position in the seed image, thus preserving the fingerprint identity (see Figure \ref{exemplosSaidaCyclegan}). Open and closed pores were also observed. We noticed other L3 traits, such as distinct ridge contours and incipient ridges, features that increase the reliability of fingerprint recognition \cite{lee2017partial, jain2010latent}. Figure~\ref{poro_open} shows examples of L3 traits present in the L3-SF database. 
 
\begin{figure}[H]
   \centering
            \begin{tabular}{cc}
            \includegraphics[scale=0.35]{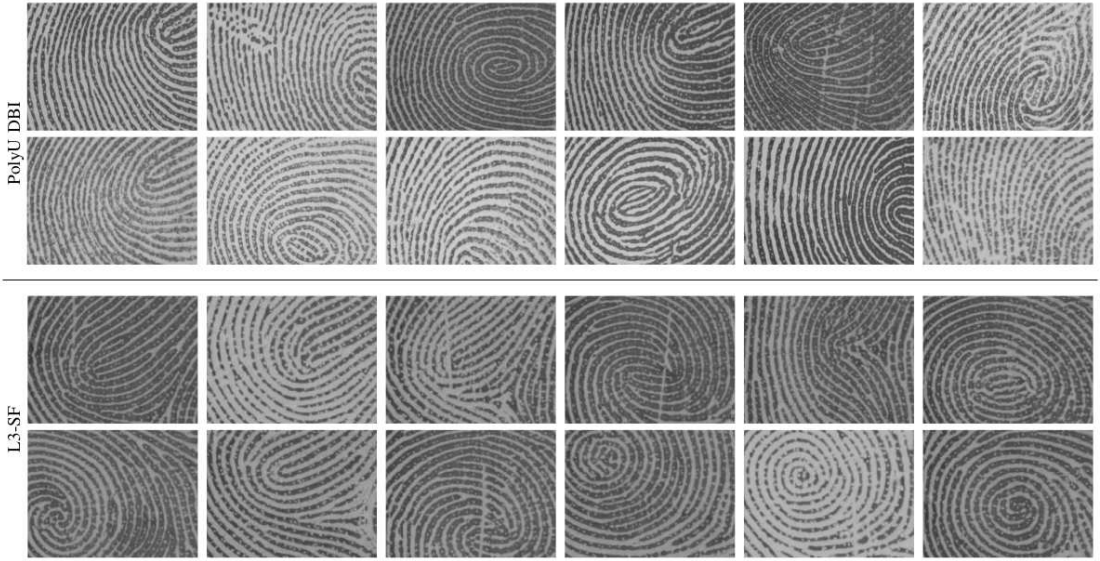} 
            \end{tabular}

  \caption{Visual comparison between real fingerprints from PolyU DBI (top) and synthetic ones from L3-SF (bottom).}
  \label{variation_background}
\end{figure}

A visual comparison between real and our synthetic fingerprints is shown in Figure~\ref{variation_background}. Note that synthetic images have different styles similarly to real ones. We selected these images randomly to provide an unbiased judgment.

\begin{figure}[H]
   \centering
            \begin{tabular}{cc}
            \includegraphics[scale=0.45]{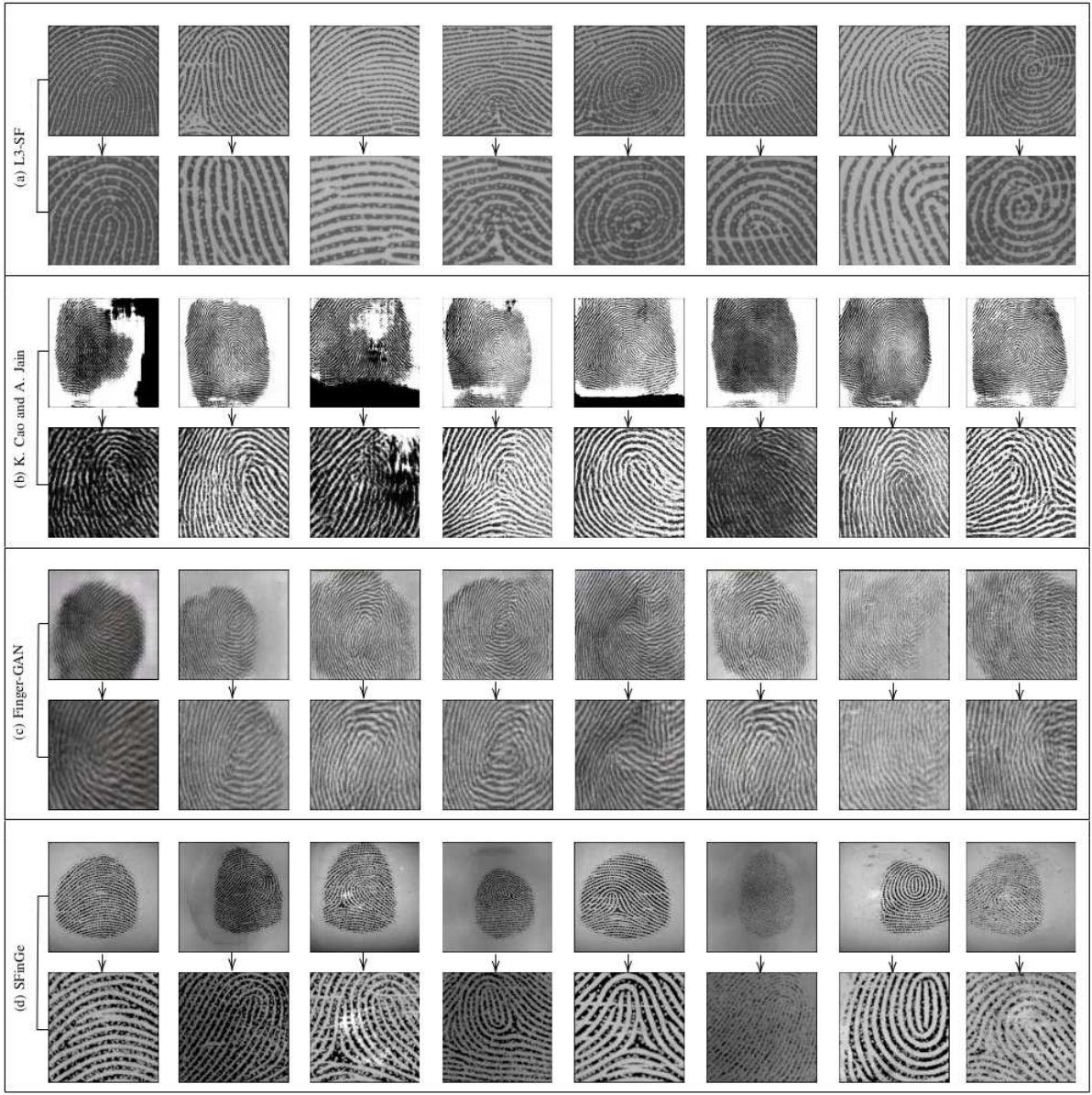} 
            \end{tabular}

\caption{Visual comparison between (a) the proposed approach, (b) Cao and Jain \cite{8411200}, (c) Finger-GAN~\cite{minaee2018fingergan} and (d) SFinGe~\cite{article_sfinge}. }
\label{compare}
\end{figure}

Figure~\ref{compare} shows a visual comparison between fingerprints generated by the proposed approach, by a publicly available SFinge demo~\cite{article_sfinge}, by a public model of Cao and Jain's method \cite{8411200}, and by our implementation of Finger-GAN~\cite{minaee2018fingergan}.

L3-SF fingerprints contain different realistic aspects, such as pores with different sizes and shapes, and ridges with acute details and texturization. Meanwhile, SFinGe~\cite{article_sfinge} generates rectangular, single-sized pores and Finger-GAN~\cite{minaee2018fingergan} does not generate pores at all. Besides, Finger-GAN and Cao and Jain's method~\cite{8411200} produce unnatural ridge shapes. Cao and Jain's method also produces irregular minutiae patterns and irregular occlusions.

\subsection{Fingerprint recognition analysis}
\label{sec:Experiments}

The goal of this experiment is to compare real and synthetic images in terms of recognition performance. Ideally, both should have close results. To carry out this comparison, first we utilize Bozorth3~\cite{bozorth}, which is a minutiae-based fingerprint matching approach. To perform pore-based fingerprint matching, we utilize Segundo and Lemes' approach~\cite{7301328}. For these analyses, we use the same protocol proposed by Liu et al.~\cite{Liu2010FingerprintPM} for PolyU DBI and all L3-SF versions, as all of them have the same configuration (148 subjects, 2 sessions, 5 images per subject per session, 320$\times$240 images). 

\begin{figure}[H]
   \centering
            \begin{tabular}{cc}
            \includegraphics[scale=0.3]{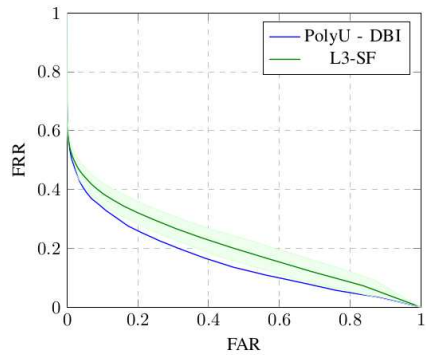} 
            \end{tabular}

  \caption{ROC curves from a minutiae-based matcher for PolyU DBI and L3-SF. For the latter, we show the average ROC and a 95\% confidence for its five subsets. }
  \label{roc}
\end{figure}

\begin{figure}[H]
   \centering
            \begin{tabular}{cc}
            \includegraphics[scale=0.3]{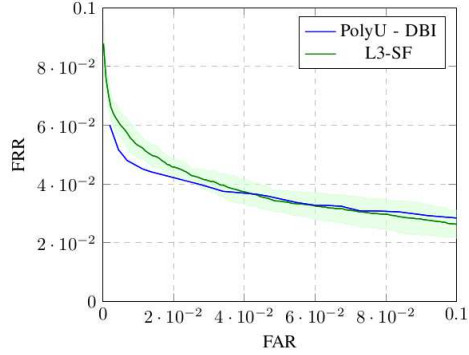} 
            \end{tabular}

  \caption{ROC curves from a pore-based matcher for PolyU DBI and L3-SF database. For the latter, we show the average ROC and a 95\% confidence for its five subsets. }
  \label{roc2}
\end{figure}

We computed the False Rejection Rate (FRR) and the False Acceptance Rate (FAR) using different threshold values for both matching approaches in all test databases. With the obtained FRR and FAR values, we plot a Receiver Operating Characteristic (ROC) curve for PolyU DBI. For our synthetic databases, we plot the average ROC with a 95\% confidence interval. These plots are shown in Figures~\ref{roc} and \ref{roc2}.

The Equal Error Rate (EER) values for minutiae-based matching are 23.84\% for real images and 28.12\% for synthetic ones. The EER values for pore-based matching are 3.37\% for real images and 3.80\% for synthetic ones. These results validate the L3-SF subsets as realistic databases, as existing recognition methods were successful without any adjustments. Besides, the accuracy in both real and synthetic datasets was very close. The larger gap in minutiae-based matching can be caused by a difference in minutiae distribution in real and synthetic images. We leave this investigation as future work.

\subsection{Human Perception Experiment}
\label{sec:humanperception}

Humans generate conscious and unconscious inferences from stimulus on the visual cortex \cite{fregnac2015cortical}. These inferences provide a judgment mechanism about different characteristics in images, where each person perceives and interprets stimuli in different ways and strategies. Based on this fact, to evaluate people's visual perception of real and synthetic images, we performed an experiment similar to the ``real vs. fake" approach popularly used on Amazon Mechanical Turk (AMT). In this experiment, we randomly show five PolyU images and five L3-SF fingerprints to the participants, which annotate the five images they consider false. Each participant repeats this task 10 times. This methodology forces the participants to identify visual characteristics and develop strategies to classify the two image types.

\begin{figure}[H]
   \centering
            \begin{tabular}{cc}
            \includegraphics[scale=0.3]{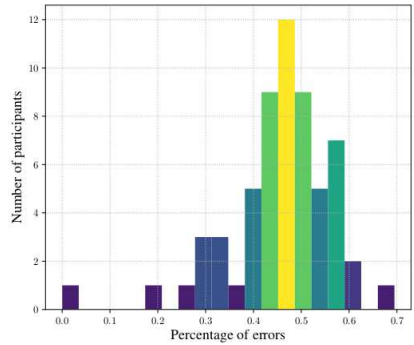} 
            \end{tabular}

  \caption{Histogram of the human perception experiment. Several participants had a classification accuracy close to random (50\%).}
  \label{Histogram}
\end{figure}

Sixty volunteers participated in this experiment, which allowed us to analyze the overall human behavior in this classification task. When participants fail to notice any visual characteristics that help discriminating real and synthetic images, their classification accuracy is close to random (50\%). Figure~\ref{Histogram} shows the histogram of human classification error. As can be seen, the average misclassification rate for all participants was 45.65\%. This result shows that the participants had difficulties distinguishing the two classes and highlights the realism of our synthetic fingerprints. 

\section{Conclusions}

We presented an approach to generate realistic, high-resolution synthetic fingerprints containing different L3 traits. We trained a CycleGAN using real fingerprint images from PolyU and handcrafted seed images to create a model capable of translating these two image domains while preserving all identification cues (e.g., ridges, minutiae, pores). Using this approach, we created the L3-SF database with the same characteristics of the PolyU DBI. More importantly, the L3-SF database allows further studies in the field of fingerprint biometrics without raising privacy-related legal issues. Our experimental results show that L3-SF images can be used by existing fingerprint recognition methods without any adjustments and achieve similar recognition performance. We performed a human perception experiment with 60 volunteers, which evidenced the realism of our synthetic images thanks to a nearly-random human classification performance. Finally, we visually compared our results with the best performing works in the literature to highlight the quality enhancement over existing works.

\bibliographystyle{unsrt}  
\bibliography{references}

\begin{thebibliography}{10}

\bibitem{Maltoni2009}
D.~Maltoni, D.~Maio, A.~K. Jain, and S.~Prabhakar.
\newblock {\em Handbook of Fingerprint Recognition}.
\newblock Springer London, 2009.

\bibitem{jain2010biometrics}
A.~K. Jain and A.~Kumar.
\newblock Biometrics of next generation: An overview.
\newblock {\em Second Generation Biometrics}, 12(1):2--3, 2010.

\bibitem{7301328}
M.~Pamplona~Segundo and R.~de~Paula~Lemes.
\newblock Pore-based ridge reconstruction for fingerprint recognition.
\newblock In {\em 2015 IEEE CVPR Workshops}, pages 128--133, June 2015.

\bibitem{DBLP:journals/corr/abs-1905-06981}
V.~Anand and V.Kanhangad.
\newblock Porenet: Cnn-based pore descriptor for high-resolution fingerprint
  recognition.
\newblock {\em CoRR}, abs/1905.06981, 2019.

\bibitem{DBLP:journals/corr/abs-1905-11472}
D.~Nguyen and Anil~K. Jain.
\newblock End-to-end pore extraction and matching in latent fingerprints: Going
  beyond minutiae.
\newblock {\em CoRR}, abs/1905.11472, 2019.

\bibitem{articleEspinoza}
M.~Espinoza and C.~Champod.
\newblock Using the number of pores on fingerprint images to detect spoofing
  attacks.
\newblock {\em 2011 International Conference on Hand-Based Biometrics, ICHB
  2011 - Proceedings}, Nov 2011.

\bibitem{inproceedingsMarcialis}
G.~L. {Marcialis}, F.~{Roli}, and A.~{Tidu}.
\newblock Analysis of fingerprint pores for vitality detection.
\newblock In {\em 2010 20th International Conference on Pattern Recognition},
  pages 1289--1292, Aug 2010.

\bibitem{inproceedingsSilva}
M.~da~Silva, A.~Marana, and A.~Paulino.
\newblock On the importance of using high resolution images, third level
  features and sequence of images for fingerprint spoof detection.
\newblock In {\em IEEE International Conference on Acoustics, Speech and Signal
  Processing}, pages 1807--1811, Apr 2015.

\bibitem{nist30}
"National~Institute of~Standards and Technology".
\newblock Nist special database 30.
\newblock {\em https://www.nist.gov/srd/nist-special-database-30}, 2002.

\bibitem{10.1007/978-3-642-01793-3_61}
Q.~Zhao, L.~Zhang, D.~Zhang, and N.~Luo.
\newblock Direct pore matching for fingerprint recognition.
\newblock In Massimo Tistarelli and Mark~S. Nixon, editors, {\em Advances in
  Biometrics}, pages 597--606. Springer Berlin Heidelberg, 2009.

\bibitem{an2019crosssensor}
V.~Anand and V.~Kanhangad.
\newblock Cross-sensor pore detection in high-resolution fingerprint images
  using unsupervised domain adaptation.
\newblock {\em arXiv preprint arXiv:1908.10701}, 2019.

\bibitem{Anand2019}
V.~Anand and V.~Kanhangad.
\newblock Pore-based indexing for fingerprints acquired using high-resolution
  sensors.
\newblock {\em Pattern Analysis and Applications}, Mar 2019.

\bibitem{butler2012naturalistic}
Daniel~J Butler, Jonas Wulff, Garrett~B Stanley, and Michael~J Black.
\newblock A naturalistic open source movie for optical flow evaluation.
\newblock In {\em European conference on computer vision}, pages 611--625.
  Springer, 2012.

\bibitem{wu2018building}
Yi~Wu, Yuxin Wu, Georgia Gkioxari, and Yuandong Tian.
\newblock Building generalizable agents with a realistic and rich 3d
  environment.
\newblock {\em arXiv preprint arXiv:1801.02209}, 2018.

\bibitem{wrenninge2018synscapes}
Magnus Wrenninge and Jonas Unger.
\newblock Synscapes: A photorealistic synthetic dataset for street scene
  parsing.
\newblock {\em arXiv preprint arXiv:1810.08705}, 2018.

\bibitem{article_sfinge}
R.~Cappelli.
\newblock Sfinge: an approach to synthetic fingerprint generation.
\newblock {\em International Workshop on Biometric Technologies}, Jan 2004.

\bibitem{ansari2011generation}
A.~H. Ansari.
\newblock Generation and storage of large synthetic fingerprint database.
\newblock {\em ME Thesis, Jul}, 2011.

\bibitem{minaee2018fingergan}
S.~Minaee and A.~Abdolrashidi.
\newblock Finger-gan: Generating realistic fingerprint images using
  connectivity imposed gan.
\newblock {\em arXiv preprint arXiv:1812.10482}, 2018.

\bibitem{8411200}
K.~{Cao} and A.~{Jain}.
\newblock Fingerprint synthesis: Evaluating fingerprint search at scale.
\newblock In {\em 2018 International Conference on Biometrics (ICB)}, 2018.

\bibitem{goodfellow2014generative}
Ian~J. Goodfellow, Jean Pouget-Abadie, Mehdi Mirza, Bing Xu, David
  Warde-Farley, Sherjil Ozair, Aaron Courville, and Yoshua Bengio.
\newblock Generative adversarial networks, 2014.

\bibitem{de2014dynamic}
Rubisley de~Paula~Lemes, Maurlcio~Pamplona Segundo, Olga~RP Bellon, and Luciano
  Silva.
\newblock Dynamic pore filtering for keypoint detection applied to newborn
  authentication.
\newblock In {\em 2014 22nd International Conference on Pattern Recognition},
  pages 1698--1703. IEEE, 2014.

\bibitem{ZhuPIE17}
J.~Zhu, T.~Park, P.~Isola, and A.~A. Efros.
\newblock Unpaired image-to-image translation using cycle-consistent
  adversarial networks.
\newblock {\em CoRR}, abs/1703.10593, 2017.

\bibitem{martijnvanmensvoort2015}
M.~van Mensvoort.
\newblock The fingerprints-world-map!
\newblock {\em
  http://fingerprints.handresearch.com/dermatoglyphics/\\fingerprints-world-map-whorls-loops-arches.htm},
  mar 2015.

\bibitem{Pattern2011}
Patrick S.~P. Wang, editor.
\newblock {\em Pattern Recognition, Machine Intelligence and Biometrics}.
\newblock Springer Berlin Heidelberg, 2011.

\bibitem{dong2016accelerating}
Chao Dong, Chen~Change Loy, and Xiaoou Tang.
\newblock Accelerating the super-resolution convolutional neural network.
\newblock In {\em European conference on computer vision}, pages 391--407.
  Springer, 2016.

\bibitem{Zhang:1984:FPA:357994.358023}
T.~Y. Zhang and C.~Y. Suen.
\newblock A fast parallel algorithm for thinning digital patterns.
\newblock {\em Commun. ACM}, 27(3):236--239, March 1984.

\bibitem{Lowe:1999:ORL:850924.851523}
D.~G. Lowe.
\newblock Object recognition from local scale-invariant features.
\newblock In {\em Proceedings of the International Conference on Computer
  Vision-Volume 2 - Volume 2}, ICCV '99, pages 1150--, Washington, DC, USA,
  1999. IEEE Computer Society.

\bibitem{Lowe2004}
D.~G. Lowe.
\newblock Distinctive image features from scale-invariant keypoints.
\newblock {\em International journal of computer vision}, 60(2):91--110, 2004.

\bibitem{Rublee:2011:OEA:2355573.2356268}
Ethan Rublee, Vincent Rabaud, Kurt Konolige, and Gary Bradski.
\newblock Orb: An efficient alternative to sift or surf.
\newblock In {\em Proceedings of the 2011 International Conference on Computer
  Vision}, pages 2564--2571, 2011.

\bibitem{yu2011asift}
Guoshen Yu and Jean-Michel Morel.
\newblock Asift: An algorithm for fully affine invariant comparison.
\newblock {\em Image Processing On Line}, 1:11--38, 2011.

\bibitem{Fischler:1981:RSC:358669.358692}
M.~A. Fischler and R.~C. Bolles.
\newblock Random sample consensus: A paradigm for model fitting with
  applications to image analysis and automated cartography.
\newblock {\em Commun. ACM}, 24(6):381--395, June 1981.

\bibitem{he2016deep}
Kaiming He, Xiangyu Zhang, Shaoqing Ren, and Jian Sun.
\newblock Deep residual learning for image recognition.
\newblock In {\em Proceedings of the IEEE conference on computer vision and
  pattern recognition}, pages 770--778, 2016.

\bibitem{taigman2016unsupervised}
Yaniv Taigman, Adam Polyak, and Lior Wolf.
\newblock Unsupervised cross-domain image generation.
\newblock {\em arXiv preprint arXiv:1611.02200}, 2016.

\bibitem{kingma2014adam}
Diederik~P Kingma and Jimmy Ba.
\newblock Adam: A method for stochastic optimization.
\newblock {\em arXiv preprint arXiv:1412.6980}, 2014.

\bibitem{1227801}
P.~Y. Simard, D.~Steinkraus, J.~C. PLATT, et~al.
\newblock Best practices for convolutional neural networks applied to visual
  document analysis.
\newblock In {\em Seventh International Conference on Document Analysis and
  Recognition, 2003. Proceedings.}, pages 958--963, Aug 2003.

\bibitem{lee2017partial}
Wonjune Lee, Sungchul Cho, Heeseung Choi, and Jaihie Kim.
\newblock Partial fingerprint matching using minutiae and ridge shape features
  for small fingerprint scanners.
\newblock {\em Expert Systems with Applications: An International Journal},
  87(C):183--198, 2017.

\bibitem{jain2010latent}
Anil~K Jain and Jianjiang Feng.
\newblock Latent fingerprint matching.
\newblock {\em IEEE Transactions on pattern analysis and machine intelligence},
  33(1):88--100, 2010.

\bibitem{bozorth}
"National~Institute of~Standards and Technology".
\newblock Nist biometric image software (nbis).
\newblock {\em
  https://www.nist.gov/services-resources/software/nist-biometric-image-software-nbis},
  2015.

\bibitem{Liu2010FingerprintPM}
F.~Liu, Q.~Zhao, L.~Zhang, and D.~Zhang.
\newblock Fingerprint pore matching based on sparse representation.
\newblock {\em 2010 20th International Conference on Pattern Recognition},
  pages 1630--1633, 2010.

\bibitem{fregnac2015cortical}
Yves Fr{\'e}gnac and Brice Bathellier.
\newblock Cortical correlates of low-level perception: from neural circuits to
  percepts.
\newblock {\em Neuron}, 88(1):110--126, 2015.

\end{thebibliography}

\end{document}